# Temporal Subtyping of Alzheimer's Disease Using Medical Conditions Preceding Alzheimer's Disease Onset in Electronic Health Records

**Zhe He, PhD[1,*], Shubo Tian, BS[1,*], Arslan Erdengasileng, MS[1], Neil Charness, PhD[1], Jiang Bian, PhD[2]**
[1]Florida State University, Tallahassee, Florida USA
[2]University of Florida, Gainesville, Florida USA

**Abstract**

*Subtyping of Alzheimer's disease (AD) can facilitate diagnosis, treatment, prognosis and disease management. It can also support the testing of new prevention and treatment strategies through clinical trials. In this study, we employed spectral clustering to cluster 29,922 AD patients in the OneFlorida Data Trust using their longitudinal EHR data of diagnosis and conditions into four subtypes. These subtypes exhibit different patterns of progression of other conditions prior to the first AD diagnosis. In addition, according to the results of various statistical tests, these subtypes are also significantly different with respect to demographics, mortality, and prescription medications after the AD diagnosis. This study could potentially facilitate early detection and personalized treatment of AD as well as data-driven generalizability assessment of clinical trials for AD.*

**Introduction**

Alzheimer's disease (AD) is a progressive neurodegenerative disorder that affects an estimated 6.2 million Americans age 65 and older in 2021. This number is likely to reach 13.8 million by 2060.[1] It is a highly heterogeneous disease that varies not only in symptoms and progression but also in the risk factors for different phenotypes. AD has been considered as associated with a wide range of risk factors including age, genetics, head injuries, vascular diseases, infections, environmental factors, and many other medical conditions such as cardiovascular disease, obesity, and diabetes.[2] Subtyping of AD considering temporal changes of patients' medical conditions prior to clinical onset of AD can be useful to facilitate understanding the heterogeneity of AD and its development (e.g., what conditions or diagnoses often precede AD onset and how they are presented across what AD subpopulations) which in turn can help improve diagnosis, treatment, prognosis and disease management.

Today, the increasing availability of large-scale datasets and development of machine learning algorithms make it possible to explore AD heterogeneity in a data-driven manner. Subtyping AD patients into homogeneous groups may lead to more differentiated disease stratification that may facilitate personalized diagnosis and prognosis.[3] Previously, most data-driven AD subtyping studies have focused on creating AD subtypes with neuroimages,[3] neuropsychological data,[4] and neuropathological data.[5,6] Electronic health records (EHRs) capture a variety of important information collected in the encounters, including demographics, diagnoses, symptoms, prescriptions, lab tests, procedures, etc. The wide adoption of EHR systems enables EHR data to be a promising source for AD subtyping. However, very few studies have used EHRs with fine-grained encounter information to identify subtypes of AD.[7] Among the handful of studies using EHR data for AD subtyping, Xu et al. identified probable AD patients in EHRs from Weill Cornell Medicine (WCM)/NewYork-Presbyterian Hospital (NYP) and then applied hierarchical clustering with variables that can predict different clinical outcomes of AD and found four subtypes of probable AD.[7] Their subtypes were derived by the clustering algorithm using the features identified in a supervised case-control-based binary prediction model. Landi et al. first employed deep learning to derive representations for AD patients in the Mount Sinai Health System Data Warehouse, and then applied hierarchical clustering to split patients with AD into three subgroups.[8]

In this study, we used the EHRs obtained from the OneFlorida Data Trust to find subtypes of AD using 40 top medical conditions diagnosed by the physicians or self-reported by the patients before first AD diagnosis among 29,922 AD patients. Our hypothesis is that occurrences and progression of the health conditions prior to AD diagnosis are correlated with different phenotypes of AD.[9] Spectral clustering algorithm was used to cluster the AD patients into subgroups based on the conditions before first AD diagnosis. We further extracted and analyzed the demographics, mortality outcome, and prescription information for the patient cohort and performed Chi-square tests and multinomial logistic regression to (1) evaluate the quality of the clustering results; and (2) identify clinically meaningful associations between the demographics, mortality, and the identified clusters. The contribution of this work is two-folds: (1) we demonstrated the feasibility of using longitudinal condition information prior to first AD diagnosis to find clinically meaningful subtypes of AD; and (2) we identified patients with certain demographics that are more likely to be clustered into certain subtypes.

---

[*] Equal-contribution first authors

**Materials and Methods**

*Data source and cohort selection*

The data used in this study were acquired from the OneFlorida Data Trust, a centralized data repository for the OneFlorida Clinical Research Consortium. OneFlorida is one of the 9 clinical data research networks in the United States funded by the Patient-Centered Outcomes Research Institute that constitute the National Patient-Centered Clinical Research Network (PCORnet).[10] The OneFlorida Data Trust, which follows the PCORnet Common Data Model, contains fine-grained and longitudinal encounter information along with diagnoses, medications, lab tests, procedures, etc. for over 15 million Floridians. We defined the patient cohort as those who were diagnosed with AD between January 2012 and January 2021 in the OneFlorida EHRs. A total of 122,669 AD patients were identified with International Classification of Disease Ninth Revision (ICD-9) code 331.0 and ICD-10 CM codes of G30, G300, G30.0, G301, G308, G30.8, G309, G30.9.

The objective of this study is to identify the subgroups of AD patients using conditions diagnosed or self-reported across 6 consecutive timeslots of 6 months each before the first diagnosis of AD. After AD patients were extracted from the OneFlorida database, a comprehensive process, as shown in Figure 1, was used to select the AD patient cohorts. We selected the patient cohort with the following selection criteria: (1) a patient should be at least 20 years old on the first AD diagnosis date, (2) has diagnosis or condition records pertaining to the top 40 prevalent conditions (using the phecode[11]) in the cohort in any of the 6 timeslots prior to the first AD diagnosis date.

*Phenotype Identification*

We used a product of the Phenome-wide association studies (PheWAS) called phecode[11] to determine the other conditions that the AD patients often also had.[12] Phecodes were introduced to present the most frequent health conditions in EHR data. Phecode groups relevant ICD-9 and ICD-10-CM codes into clinically meaningful phenotypes, thereby enabling us to leverage accumulated ICD-9 and ICD-10-CM data for PheWAS in the EHR.[12] In the OneFlorida Data Trust, ICD-9 and ICD-10 CM codes were used for records of diagnoses and self-reported health conditions. We mapped the ICD-9 and ICD-10 CM codes to phecodes and identified the most common conditions. Top 60 phenotypes were reviewed manually and the top 40 were selected as the most interested phenotypes, which are listed in Table 1.

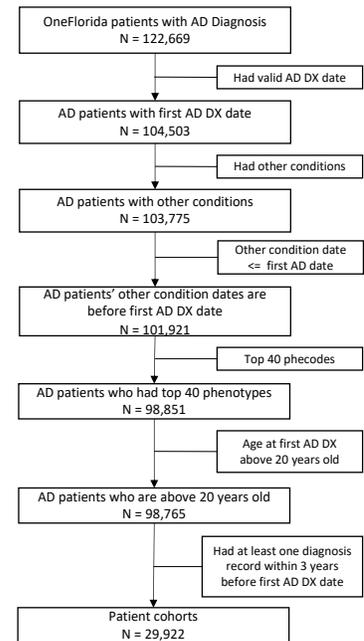

**Figure 1.** Patient cohort selection process

**Table 1.** Top 40 phecodes and phenotypes

| Phecode | Phenotype | Phecode | Phenotype |
|---|---|---|---|
| 290.1 | Dementias | 480 | Pneumonia |
| 401.1 | Essential hypertension | 740.9 | Osteoarthrosis NOS |
| 272.1 | Hyperlipidemia | 244.4 | Hypothyroidism NOS |
| 591 | Urinary tract infection | 496 | Chronic airway obstruction |
| 798 | Malaise and fatigue | 348.8 | Encephalopathy, not elsewhere classified |
| 285 | Other anemias | 428.1 | Congestive heart failure (CHF) NOS |
| 292.4 | Altered mental status | 563 | Constipation |
| 745 | Pain in joint | 427.21 | Atrial fibrillation |
| 250.2 | Type 2 diabetes | 760 | Back pain |
| 411.4 | Coronary atherosclerosis | 433.31 | Transient cerebral ischemia |
| 530.11 | GERD | 038 | Septicemia |
| 532 | Dysphagia | 741.3 | Difficulty in walking |
| 296.22 | Major depressive disorder | 276.14 | Hypopotassemia |
| 418 | Nonspecific chest pain | 509.1 | Respiratory failure |
| 785 | Abdominal pain | 585.3 | Chronic renal failure [CKD] |

| 300.1 | Anxiety disorder | 401.22 | Hypertensive chronic kidney disease |
| 350.2 | Abnormality of gait | 261.4 | Vitamin D deficiency |
| 110.11 | Dermatophytosis of nail | 296.2 | Depression |
| 585.1 | Acute renal failure | 443.9 | Peripheral vascular disease, unspecified |
| 276.5 | Hypovolemia | 295.3 | Psychosis |

To create the temporal condition information for patients, we investigated the top conditions in six consecutive 6-month timeslots in the 3 years before the first AD diagnosis for each patient, as illustrated in Figure 2. Here, Timeslot 1 is from six months before the first AD diagnosis to the first AD diagnosis; Timeslot 2 is from 12 months to seven months prior to the first AD diagnosis, so on so forth. As long as a patient was diagnosed with or self-reported a condition in a particular timeslot, we considered this patient has this condition in that timeslot.

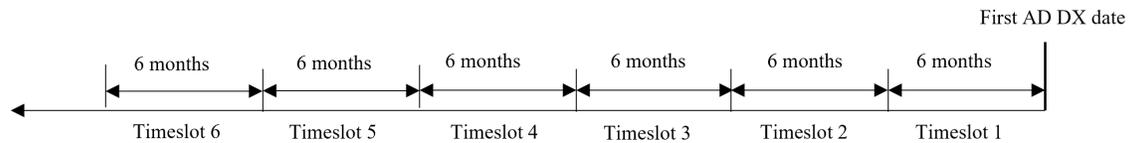

**Figure 2.** Timeslots defined for creating the temporal condition information for patients

### Clustering analysis

Clustering is a fundamental machine learning technique that uses clustering algorithms to exploit underlying structure in the data and group the data points with similar characteristics into clusters.[13] There is a wide range of clustering algorithms. The most popular ones include K-Means,[14] latent class analysis (LCA),[15] and Spectral Clustering.[16] K-Means is a widely used centroid-based clustering algorithm that groups data points into clusters by minimizing the squared distances of points to the centroid in each cluster. LCA is a popular model-based clustering approach that discovers the unobserved (or latent) classes based on the distributions of observed variables in the data with iterative, maximum likelihood method. Despite their popularity, K-Means can only separate clusters that are linearly separable in the original space, while LCA requires the assumption that the observed variables within the data are conditionally independent of each other. In recent years, spectral clustering has emerged as another popular clustering algorithm providing better performance than other algorithms in many cases, especially when the instances cannot be linearly separated in the original space. Therefore, we selected spectral clustering for this study.

Spectral Clustering is a technique originated from graph theory, but is often used on both graph data and other data. It uses the top eigenvectors of an affinity matrix derived from the data to partition the data points into clusters.[17] In this study, we used the scikit-learn Python package for the implementation of spectral clustering.[18] The affinity matrix was constructed from the Hamming distances between data points using a Laplacian kernel. K-Means was used as the strategy for assigning labels in the Laplacian embedding space.

To investigate the consistency of grouping the cohorts into different clusters, we organized the top other conditions in aggregate and temporal formats. In the aggregate format, each patient was represented by a vector of the top 40 conditions with binary values across the three-year period. In the temporal format, we used a vector of the top 40 conditions with binary values in 6 timeslots as representation of each patient. We compared the clustering results of these two approaches.

Many clustering algorithms require the number of clusters to be predefined. Various methods have been devised for choosing the number of clusters, such as Akaike information criterion (AIC), Bayesian information criterion (BIC) for model-based algorithms or Elbow method, silhouette score for non-model-based algorithms.[16] Given their simplicity and intuitive results, we used the Elbow method and K-Means to determine the number of clusters in this study. For both the clustering with aggregate conditions and temporal conditions, we ran K-Means with 1 to 10 clusters and calculated the sum of squared distances for each number of clusters. The sums of squared distances were plotted as a line against the cluster numbers. The optimal number of clusters was indicated by the inflection point on the line, i.e., the "elbow". We identified the optimal number of 4 clusters for both scenarios for this study.

### Analysis of the resulting clusters

After subgroups of the AD patient cohort were identified in the clustering analysis, statistical analyses were performed to examine the differences between the clusters with respect to (1) other conditions before the first AD diagnosis, (2) demographics, (3) mortality outcome, and (4) prescriptions after the first AD diagnosis. We also used multinomial logistic regression to model the nominal clustering results using demographic features as independent variables.

Prescription after the first diagnosis of AD can provide useful information about the treatment patterns characterizing the subgroups of AD patients. We extracted the prescription medication information after the first AD diagnosis which was available for 2,012 patients in the cohort. Prescription medication information in the OneFlorida Data Trust is coded with RxCUIs in RxNORM. To focus on the drug types instead of drugs along with dosage and dose form information, we mapped RxCUIs to Level 3 concepts of the Anatomical Therapeutic Chemical (ATC) Classification System, a drug classification system that classifies the active ingredients of drugs in five different levels where Level 3 concepts are often used to identify chemical, pharmacological or therapeutic subgroup.[19] We selected 13 most frequently prescribed drug types in terms of Level 3 ATC concepts that are relevant to AD patients and analyzed the drug types for each cluster.

**Results**

*Characteristics of the Cohort*

After the careful selection, a total of 29,922 AD patients were included in this study. The demographic information of the patient cohort in terms of sex, race and age on the first AD diagnosis date is listed in Table 2.

**Table 2.** Demographics of AD patient cohorts for clustering

|  | Patients | Percent |
|---|---|---|
| **Total** | **29,922** | **100%** |
| **Sex** | | |
| Female | 20,951 | 70.0% |
| Male | 8,971 | 30.0% |
| **Race** | | |
| American Indian or Alaska Native | 36 | 0.1% |
| Asian | 368 | 1.2% |
| Black or African American | 4,599 | 15.4% |
| Native Hawaiian or Other Pacific Islander | 7 | 0.0% |
| White | 13,065 | 43.7% |
| Multiple Race | 208 | 0.7% |
| Refuse to Answer | 10 | 0.0% |
| Other | 8,462 | 28.3% |
| Unknown | 3,167 | 10.6% |
| **Age on 1st AD Diagnosis Date (Years)** | | |
| < 65 | 2,767 | 9.2% |
| 65-75 | 5,201 | 17.4% |
| 75-85 | 10,249 | 34.3% |
| >= 85 | 11,705 | 39.1% |

*Clustering results*

We performed clustering analyses using the top 40 conditions prior to the first AD diagnosis in both the aggregate and temporal formats. We labelled the clusters resulting from the aggregate format as Cluster A, B, C, D, and the clusters resulting from the temporal format as Cluster 0, 1, 2, 3. Clustering with aggregate conditions clustered patients into cluster A, B, C, D with 10141, 5910, 6599, 7272 patients, respectively. Clustering with temporal conditions clustered patients into Cluster 0, 1, 2, 3 with 5107, 13620, 6005, 5190 patients, respectively. Cluster A (N=10,141) overlapped mostly with Cluster

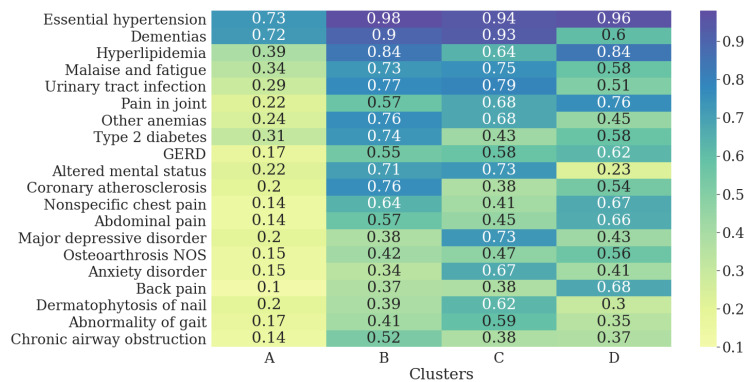

**Figure 3.** Percentage of patients having certain condition (top 20) in total patients for each cluster by clustering with aggregate conditions (n=10141, 5910, 6599, 7272 for cluster A, B, C, D)

1 (N=8,223). Cluster 1 (N=13,620) had most overlap with Cluster A (N=8,223) followed by Cluster D (N=3,154) and C (N=1,544).

To examine the characteristics of clusters with respect to conditions prior to the first AD diagnosis, we compared the percentage of patients with a certain condition in each cluster. Regarding the clustering result with aggregate conditions, as shown in Figure 3, Cluster A can be characterized as more patients diagnosed with essential hypertension and dementia but fewer patients with other conditions. Compared to Cluster A, Cluster B, C, and D had more patients with multiple conditions. Although Cluster B, C and D had more patients diagnosed with essential hypertension, they differed in other conditions with more patients. For example, Cluster B and C had more patients with dementia compared to Cluster D, while Cluster B and D had more patients with hyperlipidemia compared to cluster C. Cluster B and C had more patients with malaise, urinary tract infection, anemias, altered mental status than other two clusters.

For clustering with temporal conditions, we compared the percentage of patients with a certain condition in each timeslot for each cluster. As shown in Figure 4, the patients in Cluster 0 had a long history of essential hypertension and type 2 diabetes but only had dementia close to the first AD diagnosis. Cluster 1 and 2 had more patients with dementia only in the past 6 months prior to the first AD diagnosis. Cluster 2 and 3 had more patients with other conditions close to first AD diagnosis compared to other clusters. Cluster 3 had more patients with essential hypertension in the past 3 years and more patients with dementia in the past 2 years prior to the first AD diagnosis. A higher percentage of patients in Cluster 3 has major depressive disorder.

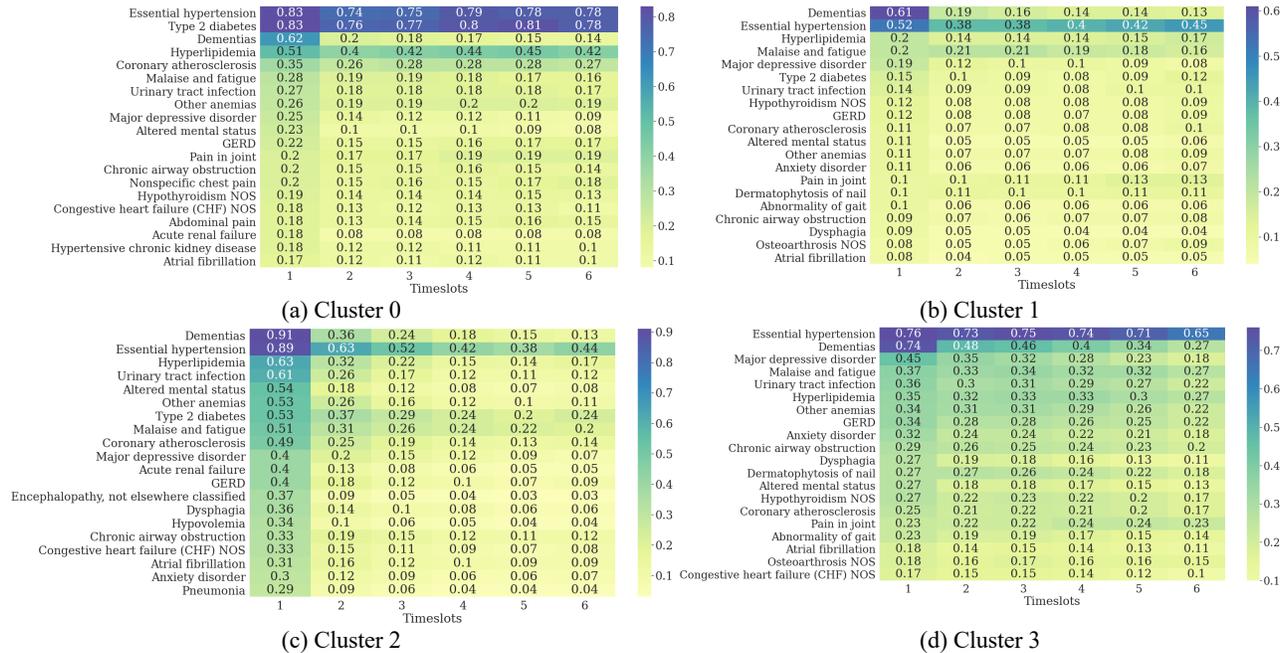

**Figure 4.** Percent of patients with certain condition (top 20) in total patients having diagnoses in each timeslot with temporal conditions in (a) Cluster 0 (N=5,107), (b) Cluster 1 (N=13,620), (c) Cluster 2 (N=6,005), and (d) Cluster 3 (N=5,190).

Comparing the results of clustering with aggregate vs. temporal conditions, we found that clustering with temporal conditions was more capable of creating subgroups that are clinically interpretable. Hence, we examined the distinct characteristics of each cluster by clustering with temporal conditions in the following subsections.

*Demographics characteristics and mortality outcome*

Figure 5 shows the characteristics of the clusters with respect to sex, race and age group on the first AD diagnosis date, as well as the mortality outcome for each cluster by clustering with temporal conditions. Cluster 3 had the highest percentage of female (N=3,815, 73.5%) and lowest percentage of male (N=1,375, 26.5%) while Cluster 2 had the lowest percentage of female (N=4,012, 66.8%) and highest percentage of male (N=1,993, 33.2%). Regarding race, Cluster 0 had the highest percentage of Black or African American (N=100, 20.0%) and lowest percentage of White (N=1,623, 31.8%), while Cluster 3 had the highest percentage of White (N=2,851, 54.9%) and lowest percentage of

Black or African American (N=687, 13.2%). Regarding age groups, cluster 0 had the highest percentage of patients in group <65 (N=605, 11.9%) and lowest percentage of patients in group >=85 (N=1,420, 27.8%) while Cluster 2 had the lowest percentage of patients in group <65 (N=388, 6.5%) and highest percentage of patients in group >=85 (N=2,553, 42.5%). According to Chi-square tests, clusters were statistically significantly different by sex (P=3.92E-15), race (P=8.76E-166) and age groups (P=1.14E-99). Regarding mortality outcome, Cluster 2 had the highest mortality rate (24.7%), followed by Cluster 3 (23.0%), Cluster 0 (17.7%), and Cluster 1 (17.6%).

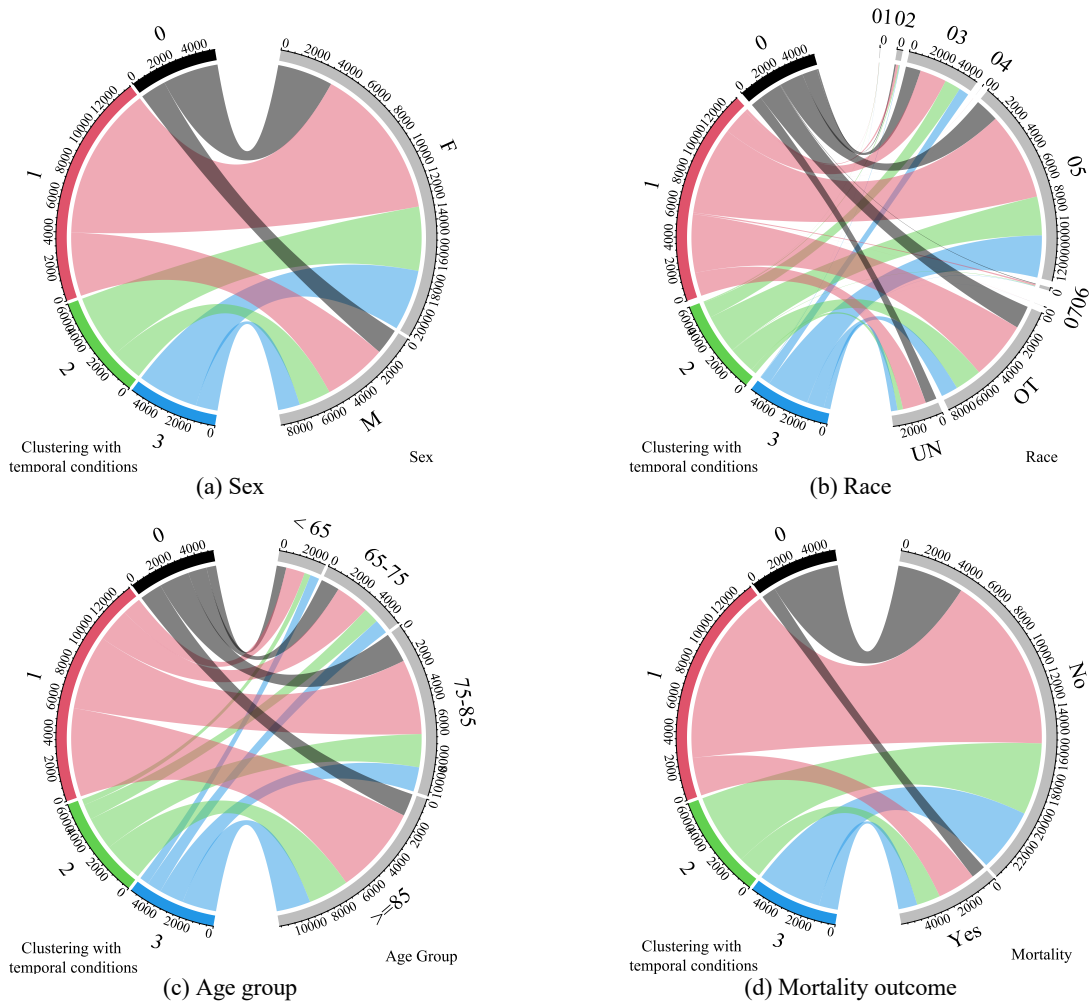

**Figure 5.** Number of patients in each cluster resulting from clustering with temporal conditions stratified by (a) sex: F-female, M-male; (b) race: 01-American Indian or Alaska Native, 02-Asian, 03-Black or African American, 04-Native Hawaiian or Other Pacific Islander, 05-White, 06-Multiple Race, 07-Refuse to Answer, OT-Other, UN-Unknown; (c) age groups on 1st AD diagnosis date; (d) mortality: No-did not die, Yes-died. (n=5107, 13620, 6005, 5190 for cluster 0, 1, 2, 3)

To examine the difference in each pair of clusters, Chi-square tests were conducted for each demographic variable and the mortality outcome. Table 3 shows the p-values of these tests. At the significance level of 0.05, all the pairs of clusters were statistically significantly different in sex except for the pair of Cluster 0 and 2 (P=0.099). All the pairs of clusters were statistically significantly different in race. The percentages of Asian and White in each pair of clusters were significantly different except for the pair of Cluster 1 and 2 (P=0.530 and 1.000), while the percentage of Black or African American in each pair of clusters were significantly different except for the pair of Cluster 1 and 3 (P=0.944). With respect to age, all the pairs of clusters were significantly different, while certain age groups were not significantly different among some pairs. For mortality, all the pairs of clusters with patients who died vs. did not die were significantly different except for the pair of Cluster 0 and 1. To ascertain the statistical importance of the subtypes, we also applied a multiple hypothesis correction method Bonferroni correction to adjust the significance level by dividing 0.05 by 15 (number of hypotheses for each pair of clusters). As such, the differences pertaining to black or African American for Cluster 0 vs. Cluster 2, multiple races for Cluster 1 vs. Cluster 3, and age group 75-85 for Cluster 0 vs. Cluster 2 (the p-values in italic) became insignificant with the new significance level of 0.003. The other conclusions made with the significance level of 0.05 remained the same.

Table 3. P-values of pairwise and all-clusters Chi-square tests for sex, race, age on 1st AD diagnosis date, and mortality outcome by clusters with temporal conditions.

|  | p-values of Chi-square tests | | | | | | |
|---|---|---|---|---|---|---|---|
|  | Pairwise by clusters | | | | | | All-clusters |
|  | 0 vs. 1 | 0 vs. 2 | 0 vs. 3 | 1 vs. 2 | 1 vs. 3 | 2 vs. 3 | |
| Sex | # | 0.099 | # | # | # | # | # |
| Race | # | # | # | # | # | # | # |
|    American Indian or Alaska Native | 0.536 | 0.053 | 0.520 | 0.077 | 1.000 | 0.288 | 0.086 |
|    Asian | # | # | # | 0.530 | # | 0.010 | # |
|    Black or African American | # | *0.007* | # | # | 0.944 | # | # |
|    Native Hawaiian or Other Pacific Islander | 0.901 | 0.935 | 1.000 | 0.317 | 0.887 | 0.942 | 0.474 |
|    White | # | # | # | 1.000 | # | # | # |
|    Multiple Races | 0.428 | 0.082 | 0.003 | 0.223 | *0.008* | 0.190 | 0.010 |
|    Refuse to Answer | 0.728 | 0.935 | 0.628 | 0.237 | 0.990 | 0.199 | 0.298 |
|    Other | 0.114 | 0.895 | # | 0.137 | # | # | # |
|    Unknown | # | # | # | # | # | # | # |
| Age groups (age on 1st AD diagnosis date, years) | # | # | # | # | # | # | # |
|    < 65 | # | # | 0.300 | # | # | # | # |
|    65-75 | # | # | # | 0.305 | 0.002 | 0.071 | # |
|    75-85 | # | *0.009* | # | 0.911 | # | # | # |
|    >= 85 | # | # | # | 0.039 | 0.424 | 0.327 | # |
| Mortality | 0.802 | # | # | # | # | 0.034 | # |

Note: #: p-values <= 0.001. n=5107, 13620, 6005, 5190 for cluster 0, 1, 2, 3.

*Prescription*

Prescription information is only available for 2,012 patients in this cohort. Figure 6 shows the percent of patients prescribed with selected drugs in each cluster by clustering with temporal conditions. All the clusters had a higher percentage of patients prescribed with other analgesics and antipyretics than other medications. Cluster 2 and 3 had higher percentage of patients prescribed with drugs for constipation. Cluster 0 had the highest percentage of patients prescribed with drugs for peptic ulcer and GORD whereas Cluster 0 and 2 had higher percentage of patients prescribed with insulins and analogues. Note that some medications (e.g., drugs for constipation, GORD) could have been prescribed due to side effects of medications for other conditions such as diabetes and hypertension.

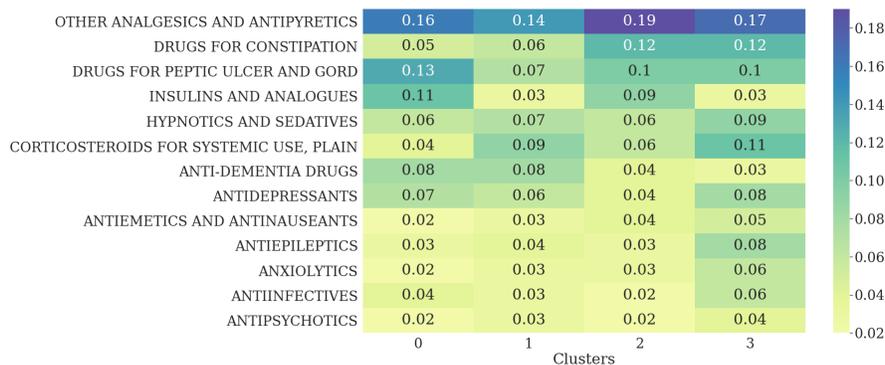

Figure 6. Percent of patients having prescription of selected drugs in total patients having prescription in each cluster by clustering with temporal conditions (n= 289, 1094, 363, 266 for cluster 0, 1, 2, 3)

*Multinomial logistic regression*

To further evaluate the clustering results and identifying the factors that cause patients to be more likely in a certain cluster, we conducted multinomial logistic regression for assigning each patient into different clusters using demographic features as independent variables (predictors). We used Cluster 0 as the reference group. According to the results in Table 4, male patients were more likely to be in Cluster 2 (Relative Risk Ratio [RRR]=1.237, p<0.01) but less likely to be in Cluster 3 (RRR=0.858, p<0.01). In terms of race, Asian patients had lower probability of being in Cluster 1 (RRR=0.326, p<0.1), Cluster 2 (RRR=0.137, p<0.01), and Cluster 3 (RRR=0.152, p<0.05). Black or African Americans and Native Hawaiian or Other Pacific Islanders were less likely to be in Cluster 2 (RRR=0.240, p<0.05; RRR=0.00002, p<0.01), whereas patients of multiple races had lower probability of being assigned into Cluster 2 (RRR=0.162, p<0.01) and Cluster 3 (RRR=0.210, p<0.05). With respect to age, patients of age 65-75 had higher chance of being in Cluster 2 (RRR=1.403, p<0.01) but had lower chance of being in Cluster 3 (RRR=0.864, p<0.05), patients of age 75-85 had higher chance of being in Cluster 1 (RRR=1.319, p<0.01) and Cluster 2 (RRR=1.912, p<0.01), whereas patients of age >=85 had higher chance of being in Cluster 1 (RRR=2.027, p<0.01), Cluster 3 (RRR=1.632, p<0.01), and Cluster 2 (RRR=3.059, p<0.01) in particular.

**Table 4.** Results of multinomial logistic regression models for predicting patients to be in different clusters by clustering with temporal conditions

| Demographic Characteristics | Cluster 1 n=13,620 | Cluster 2 n=6,005 | Cluster 3 n=5,190 |
|---|---|---|---|
| | Relative Risk Ratio (RRR) (Robust SE) | | |
| **Sex[a]** | | | |
| Male | 1.003 (0.036) | 1.237*** (0.042) | 0.858*** (0.045) |
| **Race[b]** | | | |
| Asian | 0.326* (0.651) | 0.137*** (0.663) | 0.152** (0.737) |
| Black or African American | 0.369 (0.639) | 0.240** (0.645) | 0.321 (0.710) |
| Native Hawaiian or Other Pacific Islander | 1.048 (1.270) | 0.00002*** (0.001) | 0.526 (1.584) |
| White | 0.738 (0.639) | 0.354 (0.644) | 0.809 (0.709) |
| Multiple Race | 0.445 (0.663) | 0.162*** (0.681) | 0.210** (0.756) |
| Refuse to Answer | 1.384 (1.257) | 0.00004*** (0.001) | 1.476 (1.356) |
| Other | 0.49 (0.639) | 0.239** (0.644) | 0.331 (0.710) |
| Unknown | 0.438 (0.640) | 0.110*** (0.646) | 0.286* (0.711) |
| **Age Group[c]** | | | |
| 65-75 | 0.952 (0.062) | 1.403*** (0.079) | 0.864** (0.074) |
| 75-85 | 1.319*** (0.058) | 1.912*** (0.073) | 0.926 (0.069) |
| 85+ | 2.027*** (0.059) | 3.059*** (0.075) | 1.632*** (0.069) |
| Constant | 3.578** (0.640) | 2.159 (0.647) | 2.014 (0.711) |
| AIC | 75,632.55 | | |

Note: *p<0.1, **p<0.05, ***p<0.01. Cluster #0 serves as the reference group (n=5,107). [a]Reference is Female. [b]Reference is American Indian or Alaska Native. [c]Reference is Age Group <65.

**Discussion**

Identifying meaningful subtypes of AD can support the development of prevention and treatment strategies for AD by informing the design of clinical trials as well as the analysis of their generalizability, which is often overlooked.[20] An intervention can be tested in different subtypes, for which efficacy and safety of the intervention can be evaluated separately. With state-of-the-art trial generalizability analysis methods,[21] one can also assess the generalizability of a trial using different target populations based on the subtypes, making it easier to identify the subgroups to which the results are mostly applicable.

Based on our hypothesis that medical conditions prior to AD diagnosis are associated with different subtypes of AD, we performed spectral clustering to cluster the AD cohort of 29,922 patients using their physician-diagnosed or self-reported conditions in 6 consecutive half-year periods (timeslots) before the first diagnosis of AD from EHRs and identified four subtypes of AD (subtypes 0, 1, 2, and 3). Chi-square tests indicated that the subtypes found in this study were statistically significantly different with respect to demographics, mortality, and prescriptions after AD diagnosis.

Among the 4 derived subtypes of AD, Subtype 0 (Cluster 0) can be characterized as having a long history of essential hypertension, type 2 diabetes and hyperlipidemia, short history of dementia, and fewer other conditions before the diagnosis of AD. Subtype 0 has a higher percentage of Black or African Americans and younger patients (age<75), lower mortality rate, and a higher percentage of patients prescribed with drugs for peptic ulcer and GORD and insulins. Subtype 1 can be characterized as having long history of essential hypertension, short history of dementia, and few other conditions before first AD diagnosis. Subtype 1 (Cluster 1) has a higher percentage of White, lower mortality rate, and higher percentage of patients prescribed with corticosteroids, and anti-dementia drugs. Subtype 2 (Cluster 2) can be described as having a long history of essential hypertension, short history of dementia, and multiple other conditions close to the diagnosis of AD, a lower percentage of female, a higher percentage of White and older patients (age>=75), higher mortality rate, and a higher percentage of patients prescribed with drugs for constipation, and insulins and analogues. Subtype 3 (Cluster 3) has a long history of essential hypertension and diverse history of multiple other conditions before first AD diagnosis, a higher percentage of female and White, lower mortality rate, and a higher percentage of patients prescribed with drugs for constipation, hypnotics and sedatives, corticosteroids, and antiepileptics. As this is the first study on the temporal subtyping of Alzheimer's disease using the conditions before the first AD diagnosis in EHR, the results are not directly comparable to other studies that used EHR data for AD subtyping.[7,8] Nonetheless, some clusters do share similar characteristics with previous studies. For example, patients in Subtype 2 are older (mean age: 81.9) and have more comorbidities, similar to Subphenotype C in Xu et al.[7]

There are a few limitations of this study. First, the cohort from the OneFlorida EHRs has been greatly narrowed down for our analysis (122,669->29,922) due to our focus on temporal changes of one's medical conditions in the past 3 years prior to the first AD diagnosis. Therefore, it may not be representative of the general AD population. Carrying out these analyses in other AD data sets would speak to generalizability of these results. Nonetheless, as Florida is one of the most populous states for older adults, OneFlorida Data Trust is still one of the largest EHR datasets available for such a study. Second, the information about medical conditions prior to the first AD diagnosis may be incomplete and OneFlorida contributing sites may use different diagnostic criteria and coding conventions, hence the clusters based on time interval analyses may lack sensitivity for identifying other precursors of AD. Third, OneFlorida Data Trust does not have AD progression information such as Mini-Mental State Exam (MMSE), neuroimages, or neuropathological, biochemical biomarkers which can also be used for subtyping. Nontheless, the results of this study have demonstrated the potential of using longitudinal EHR data for AD subtyping. One could also evaluate the hazard ratio for mortality for different subtypes when recruiting patients into clinical trials. Aside from the information used in this study, EHRs have rich additional information contained in unstructured text of clinical notes. Extracting information from the clinical notes with state-of-the-art natural language processing (NLP) techniques will potentially facilitate the use of EHR data for AD subtyping. In addition, deep learning has achieved great success in many applications in computer vision and NLP with multiple data modalities. Given the diversity of data in the medical domain, promising achievements can be expected for further research of multi-modal subtyping and deep phenotyping using EHR data coupled with neuroimages, neuropsychological data, and neuropathological, clinical and biochemical biomarkers.

**Conclusion**

In this study, we performed spectral clustering of AD patients using the longitudinal medical condition information before the AD diagnosis. Our analysis using OneFlorida Data Trust indicates that subtypes of AD created by this technique are significantly different in terms of the conditions, demographics, mortality, and prescription medications. Future research is warranted to further evaluate the clinically meaningfulness of these clusters. This work could facilitate not only early detection and personalized treatment of AD but also data-driven generalizability assessment of clinical trials for AD.

**Acknowledgements**

This study was partially supported by the National Institute on Aging under Award Number R21AG061431; and in part by Florida State University-University of Florida Clinical and Translational Science Award funded by National Center for Advancing Translational Sciences under Award Number UL1TR001427. The content is solely the